\documentclass[10pt,twocolumn,letterpaper]{article}

\usepackage{cvpr}
\usepackage{times}
\usepackage{epsfig}
\usepackage{graphicx}
\usepackage{amsmath}
\usepackage{amssymb}
\usepackage{float}
\usepackage{multirow,makecell,footmisc,url}
\usepackage{gensymb}
\usepackage{color}

\usepackage[pagebackref=true,breaklinks=true,letterpaper=true,colorlinks,bookmarks=false]{hyperref}

\newcommand{\norm}[1]{\left\lVert#1\right\rVert}

\cvprfinalcopy 

\def\cvprPaperID{571} 

\ifcvprfinal\fi
\begin{document}


\title{Fast Video Object Segmentation with Temporal Aggregation Network
and \\ Dynamic Template Matching }
\author{
\begin{tabular}{c@{\hspace{1cm}}c@{\hspace{1cm}}c@{\hspace{1cm}}c}
Xuhua Huang$^{1*}$ & ~~~Jiarui Xu$^{1}$\thanks{Equal contribution. This
research is supported in part by Tencent and the Research Grant Council of
the Hong Kong SAR under grant no. 1620818.}~~~ & Yu-Wing Tai$^{2}$ &
Chi-Keung Tang$^{1}$
\end{tabular}
\\
\begin{tabular}{cc}
$^{1}$The Hong Kong University of Science and Technology& $^{2}$Tencent\\
\end{tabular}
\\
\begin{tabular}{cccc}
{\tt\small xhuangat@ust.hk} & {\tt\small  jxuat@ust.hk} & {\tt\small
yuwingtai@tencent.com} & {\tt\small cktang@cs.ust.hk}\\
\end{tabular}}

\maketitle

\begin{abstract}
\vspace{-10pt}
Significant progress has  been made in Video Object Segmentation (VOS), the video object tracking task in its  finest level.
While the VOS task can be naturally decoupled into image semantic segmentation and video object tracking, significantly much more research effort has been made in segmentation than tracking.
In this paper, we introduce ``tracking-by-detection"  into VOS which can coherently integrates segmentation into tracking, by proposing a new temporal aggregation network and a novel dynamic time-evolving template matching mechanism to achieve significantly improved performance. Notably, our method is entirely online and thus suitable for one-shot learning, and our end-to-end trainable model allows multiple object segmentation in one forward pass.
We achieve new state-of-the-art performance on the DAVIS benchmark without complicated bells and whistles in both speed and accuracy, with a speed of $0.14$ second per frame and $\mathcal{J} \& \mathcal{F}$ measure of $75.9\%$ respectively. Project page is available at \url{https://xuhuaking.github.io/Fast-VOS-DTTM-TAN/}.
\end{abstract}

\begin{figure}[t]
\begin{center}
   \includegraphics[width=1.0\linewidth]{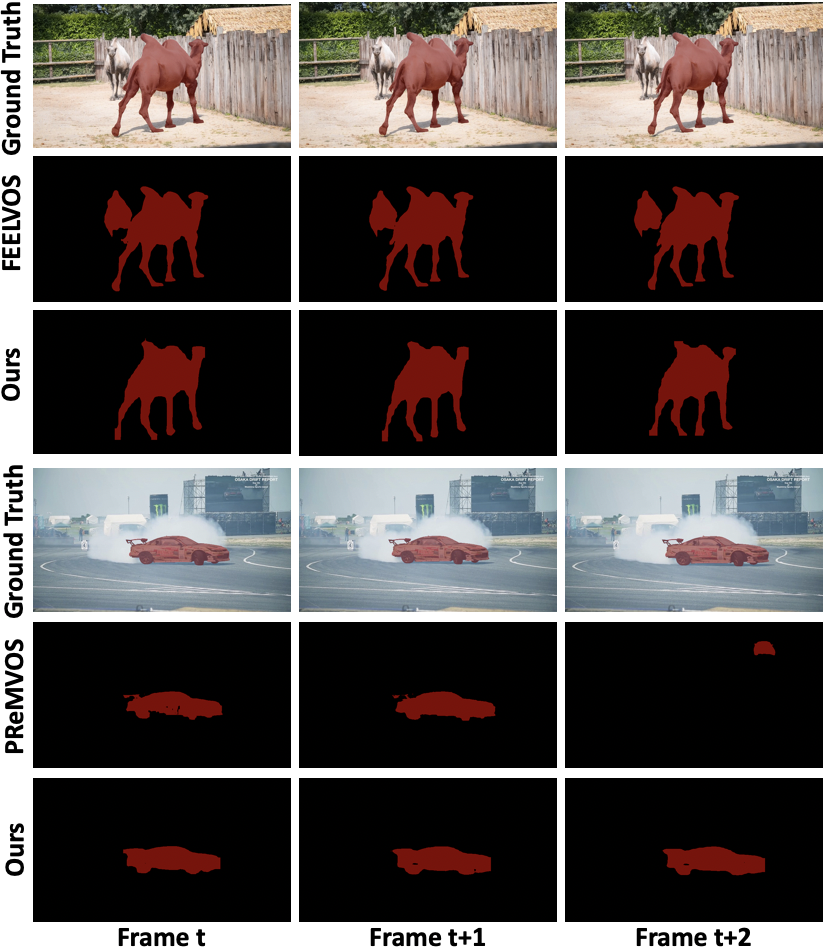}
\end{center}
   \caption{\textbf{Qualitative comparisons} on the DAVIS validation set. The camel example shows the case when FEELVOS~\cite{voigtlaender2019feelvos} misjudges pixels from another object.
   The racing car example shows that PReMVOS~\cite{luiten2018premvos} switches to another object between two consecutive frames.
   Both FEELVOS and PReMVOS are respectively the top 2 entries on the DAVIS benchmark.
   }
   \vspace*{-15pt}
\label{fig:teaser}
\end{figure}

\vspace{-15pt}
\section{Introduction}

Video object segmentation (VOS) is a fine-grained labeling problem aiming to find pixel-level correspondence across frames in a given video, which has broad range of applications including surveillance, video editing, robotics and autonomous driving.
In this work, we focus on the semi-supervised VOS setting in which the ground-truth segmentation masks of the target objects in the first frame are given. The task is then to automatically predict the segmentation masks of the target objects for the rest of the video.
With the recent advances in deep learning and the introduction of the DAVIS datasets~\cite{perazzi2016davis, pont2017davis}, tremendous progress has been made in tackling this semi-supervised VOS task.
Notwithstanding, existing state-of-the-art methods are heavily biased toward semantic segmentation in their design and thus they do not leverage the advantages of excellent tracking solutions. We believe that the VOS task can be naturally decoupled into image semantic segmentation and video object tracking:

In semantic segmentation, most works~\cite{chen2014deeplabv1, chen2017deeplabv2, chen2017deeplabv3, chen2018deeplabv3plus} are mainly based on Fully Convolutional Networks~\cite{long2015fcn}.
Many VOS pipelines~\cite{khoreva2017lucid, caelles2017osvos, voigtlaender2017online, bao2018cnn, voigtlaender2019feelvos, wang2019ranet, oh2018fast} exploit these architectures to produce  a segmentation map.
However, semantic segmentation is not instance sensitive. In complex scenarios where many different instances share the same semantics (e.g., pedestrians, vehicles and cases in Figure~\ref{fig:teaser}), these methods based on semantic segmentation may fail to track individual objects consistently.
This has prompted us to revisit the whole pipeline from the perspective of Multiple Object Tracking (MOT).

In MOT, most recent approaches~\cite{xiang2015learning, leal2016learning,  sadeghian2017tracking, chu2017online, sadeghian2017tracking, chu2017online, bochinski2017high, zhu2018online, xu2019spatial, bergmann2019tracking} have adopted the popular “tracking-by-detection” model, where objects are first localized in each frame and then associated across frames.
This model benefits from the rapid progress in the field of object detection~\cite{ross2010dpm, ren2015faster, he2017mask, lin2017feature, lin2017focal, dai2017deformable, cai2018cascade}, and has led to several popular benchmarks over the past few years, i.e., MOT15$\sim$17 \cite{MOTChallenge2015, MOT16}.
Such a decoupled pipeline also makes it easy to extend or upgrade the tracking system with latest detection techniques.

Our work is motivated by the recently proposed tracker~\cite{bergmann2019tracking}
which consists of further extension of the ``tracking-by-detection" model without exploiting tracking annotation.
Under the semi-supervised setting, even though the first frame ground-truth is given, tracking annotation is still unavailable.
We fine-tune the detector with the first frame ground-truth segmentation mask for domain adaptation, whose iterations are proportional to the length of the sequence, and then perform tracking through the rest of the video.
We further extend the tracker with our novel Temporal Aggregation Network and Dynamic Time-evolving Template Matching mechanism, both of which make the tracking pipeline more robust to occlusion and appearance changes.

Many leading methods rely on various overly-engineered design and/or heavy modules to improve their system performance while sacrificing run times.
For example, PReMVOS~\cite{luiten2018premvos}, the winner of 2018 DAVIS challenge and current champion on DAVIS benchmark, adopts a total of four neural networks, optical flow and merging mechanism, and as a result it takes around 38 seconds to segment one single frame.
DyeNet~\cite{xiao2018jointreid}, another leading method in VOS, integrates FlowNet~\cite{flownet2} for utilizing optical flow information and Bi-directional Recurrent Neural Network (RNN) for mask propagation, leading to a speed of around 2.4 seconds per frame.
Though some of these methods can generate excellent results, the long run times and/or high complexity of their system  prevent them from practical deployment or extension with advancement of relevant techniques.

In this paper, 
we present a simple, fast and high performance new baseline for VOS.
Instead of formulating video object segmentation as mainly a segmentation problem, we propose to tackle the problem as a tracking problem with segmentation in fine-grained level.
The proposed novel Temporal Aggregation Network and Dynamic Time-evolving Template Matching can be easily integrated into this simple pipeline to  boost  performance.

Our \textbf{contributions} can be summarized as follows:
\vspace{-3pt}
\begin{itemize}
   \item We present a simple, easy-to-extend and end-to-end trainable pipeline for VOS task by introducing the ``tracking-by-detection" model into VOS which supports multiple object segmentation in one forward pass. To the best of our knowledge, we are the first to introduce ``tracking-by-detection" model into video object segmentation as an online method with strong results and high speed;
\vspace{-3pt}
   \item We contribute a novel Temporal Aggregation Network and Dynamic Time-evolving Template Matching mechanism, which are entirely online and naturally fit into one-shot learning to boost performance;
\vspace{-3pt}
   \item We achieve a new state-of-the-art performance in both speed and accuracy without complicated bells and whistles on the DAVIS benchmark, with a speed of $0.14$ second per frame and $\mathcal{J} \& \mathcal{F}$ mean score of 75.9\%.
\end{itemize}

\begin{figure*}
\begin{center}
   \includegraphics[width=0.9\linewidth]{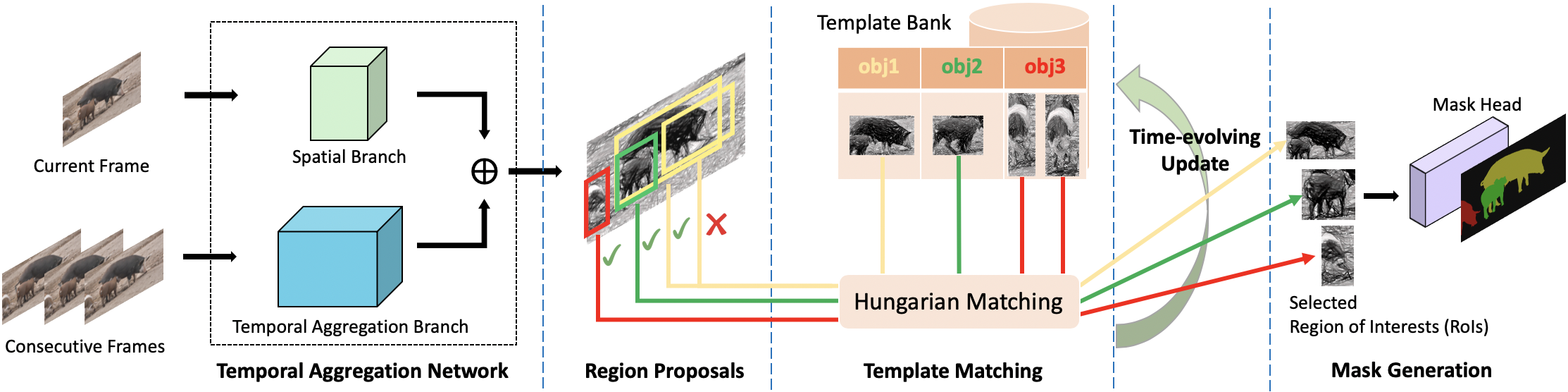}
\end{center}
\vspace{-0.2in}
   \caption{Overview of our pipeline.}
\label{fig:pipeline}
\vspace{-0.2in}
\end{figure*}

\section{Related Work}

\textbf{Semantic Segmentation}
Recently, state-of-the-art semantic segmentation frameworks based on the fully convolutional network (FCN) \cite{long2015fcn} have made remarkable progress.
Deeplab~\cite{chen2014deeplabv1, chen2017deeplabv2, chen2017deeplabv3, chen2018deeplabv3plus} and context based methods~\cite{zhao2017pyramid, wang2018non, fu2019dual, zhao2018psanet, huang2018ccnet} have further improved the performance over the years.
However, benchmarks in semantic segmentation~\cite{Cordts2016Cityscapes, zhou2017scene} are used to evaluate full image semantic labeling task, while VOS belongs to instance segmentation of one or multiple specific target objects.
For semantic segmentation, pixels in the same semantics are labeled with same category with no instance information.
As for instance segmentation, the combination of both instance detection and pixel-level segmentation are required.
The misalignment of these two tasks indicates that the design and method in semantic segmentation may not directly applicable in video object segmentation. Notwithstanding,
with this intrinsic difference, many leading entries~\cite{khoreva2017lucid, caelles2017osvos, voigtlaender2017online, bao2018cnn, voigtlaender2019feelvos, wang2019ranet, oh2018fast} in DAVIS benchmark still utilize this pipeline and adopt the  architecture of semantic segmentation, which as a result make different instances vulnerable to switch and/or drift as shown in Figure~\ref{fig:teaser}.

\textbf{Tracking-by-Detection Model}
Recent multiple object tracking (MOT) methods are mostly based on the tracking-by-detection model.
As object detection has been well studied on its own, the main focus of MOT is on the data association problem. In MOT Benchmark~\cite{MOTChallenge2015, MOT16}, public detectors are shared among all leading entries~\cite{sadeghian2017tracking, xu2019spatial}.
This makes all pipelines designs adopting this model easy to extend with the latest detection approaches~\cite{ross2010dpm, ren2015faster, he2017mask, lin2017feature, lin2017focal, dai2017deformable, cai2018cascade}.
Even though MOT and VOS Benchmark settings differ in many aspects, the same intrinsic key is robust tracking ability.
However, surprisingly only a few approaches formulate the VOS problem using this model and they do not come with online setting. While
PReMVOS~\cite{luiten2018premvos} may be regarded as some kind of close proximate to the tracking-by-detection model, its design is very complex and heavy, involving optical flow, multiple networks, and the framework cannot be trained end-to-end. 
Our method adopts the tracking-by-detection model which explicitly introduces detection and association into video object segmentation.
Compared with PReMVOS, our approach is suitable for online setting, easy to extend and end-to-end trainable.

\textbf{Video Object Segmentation}
Video Object Segmentation (VOS) aims for joint segmentation and tracking. The recently released DAVIS benchmarks~\cite{perazzi2016davis, pont2017davis, caelles2018davis} have contributed significantly in pushing the frontier of the relevant state-of-the-arts. However, as noted in~\cite{voigtlaender2019feelvos}, many of the methods do not fulfill the design goals of being robust, simple, fast and end-to-end trainable.
For instance, as the winning entry of DAVIS 2018 Challenge~\cite{caelles2018davis}, PReMVOS~\cite{luiten2018premvos} exploits 4 different neural networks working in tandem, optical flow wrapping on image space and complicated merging algorithm, taking more than 38s for processing a given frame.
The second runner up CINM~\cite{bao2018cnn} uses Markov Random Field (MRF) and iterative inference algorithm.
DyeNet~\cite{xiao2018jointreid} incorporates template matching into a re-identification network; however optical flow warping and RNN are used in the mask network, thus making it training complicated and computationally demanding.
Recently proposed FEELVOS~\cite{voigtlaender2019feelvos} is an extension of MaskTrack~\cite{perazzi2017masktrack} which uses previous frames predictions as input for the current frame,
which is prone to accumulation error during tracking.
Moreover, due to the absence of shared design model, many methods have limited practical usage and are not easy to extend.
Our introduction and extension of ``tracking-by-detection" model aims to establish a new, simple and strong baseline for video object segmentation.

\section{Method}

Refer to Figure~\ref{fig:pipeline}.
The feature map of the current frame is extracted through our Temporal Aggregation Network (TAN) which coherently combines spatial and temporal information. TAN aggregates neighboring frame spatial features across time to enrich the current frame feature.
On top of the aggregated feature map from TAN, a Region Proposal Network (RPN) is applied to extract region proposals which are likely to contain target objects.
With region proposals in the current frame, Hungarian matching~\cite{munkres1957algorithms} is performed between template features in the bank and current detected object features. Initially, the template features are just the features of the first frame ground truth objects.
After matching, the most confident region proposal for each object will be chosen and passed to the Mask Head for final segmentation generation. At the same time, if large appearance change occurs within the chosen proposals, the new proposals would serve as new templates for future frames.

We first introduce how we incorporate tracking-by-detection model into the video object segmentation task.
Then we will present our proposed Temporal Aggregation Network and Dynamic Time-evolving Template Matching.

\subsection{Tracking by Detection}
\label{section: tbd}
The goal of video object segmentation (VOS) is similar to multi-object tracking (MOT),
which is to predict the trajectories of multiple objects at a fine-grained level over time.
We denote the set of trajectories ${\bf{T}} = {\left\{ {{{\bf{T}}_i}} \right\}_{i = 1}^N}$, where
the trajectory of the $i^\text{th}$ object can be represented by a series of bounding boxes together with the masks within, denoted by ${{\bf{T}}_i} = {\left\{ {{\bf{b}}_i^t, \, {\bf{m}}_i^t} \right\}_{t = 1}^T}$, ${\bf{b}}_i^t = \left[ {x_i^t,y_i^t,w_i^t,h_i^t} \right]$; $x_i^t$ and $y_i^t$ denote the center location of the target $i$ at frame $t$; $w_i^t$ and $h_i^t$ denote respectively the width and height of the target object $i$;
${{\bf{m}}_i^t}$ denotes the corresponding foreground segmentation mask within corresponding bounding box.
In MOT usually only ${\{{{\bf{b}}_i^t}\}_{t=1}^T}$ is needed, while VOS is more fine-grained and should produce ${\{{{\bf{m}}_i^t}\}_{t=1}^T}$.
Bounding box representation is popular in recent detection pipelines \cite{ren2015faster, he2017mask}
while some non-box based methods~\cite{chen2019tensormask} have also been proposed. We mainly adopt box based detection pipelines in this paper but note that our method is also compatible with non-box based pipelines.

To output segmentation mask, a light-weight Fully Convolutional Network is attached to the detector; foreground/background segmentation is performed on each detected bounding boxes.
For simplicity, we ignore the ${\{{{\bf{m}}_i^t}\}_{t=1}^T}$ which can be easily generated from ${\{{{\bf{b}}_i^t}\}_{t=1}^T}$.
Under semi-supervised setting of DAVIS Benchmark~\cite{perazzi2016davis, pont2017davis, caelles2018davis},  the tracking targets are given in the first frame.
We denote  tracking targets as ${\{{\bf{t}}_i\}}_{i=1}^{N}$.


Our method follows the online tracking-by-detection model~\cite{xiang2015learning}, which first detects multiple objects in each frame and then associates their identities across frames.
Following~\cite{caelles2017osvos, khoreva2017lucid} we first train the whole detector offline with the training set in order to adapt the detector to the DAVIS dataset domain.
Before inference, one-shot learning is performed with the given ground-truth segmentation for a few iterations to reduce false alarm.
Note that one advantage of our training strategy is that no tracking-specific training is required, which implies that our method can be readily extended to various applications.

During inference, different from~\cite{bergmann2019tracking} and for simplicity we do  not exploit bounding box regressor for tracking.
Instead, simple Intersection over Union (IoU) metric is adopted.
For segmentation based approaches, IoU is hard to be incorporated into the pipeline since there is no explicit bounding box or instance.
Our method first detects multiple objects in each frame and then associates their identities across frames.
The detected boxes at frame $t$ with confidence score greater than $\sigma_{\text{det}}$ are denoted as $\{{\bf{b}}^t_{j}\}_{j=1}^{N_t} = D_t$.
Note that $N_t$ is not necessarily equal to $N$ due to false positives and false negatives output of the detector.
At $t=0$, our tracker initializes the tracklet from the first frame bounding boxes ${\{{\bf{t}}_i\}}_{i=1}^{N} = \{{\bf{b}}^0_{j}\}_{j=1}^{N_0} = D_0$.
At frame $t$, a bipartite graph is constructed based on location similarity IoU, the weight $\omega^{\text{loc}}_{ij}$ between $i$th target and $j$th detected object in the current frame is defined as the IoU of ${\bf{t}}_i$ and ${\bf{b}}^t_j$.
After Hungarian matching~\cite{munkres1957algorithms} is performed, the target ${\bf{t}_i}$ will be updated with ${\bf{b}}^t_k$ and added to ${\bf{T}}_i$ if matched.
When running the assignment process in a frame-by-frame manner the object trajectories are produced.

This naive tracking-by-detection pipeline under the setting of semi-supervised video object segmentation will be extended in following sections.

\subsection{Temporal Aggregation Network}
\begin{figure}[]
\begin{center}
   \includegraphics[width=1.0\linewidth]{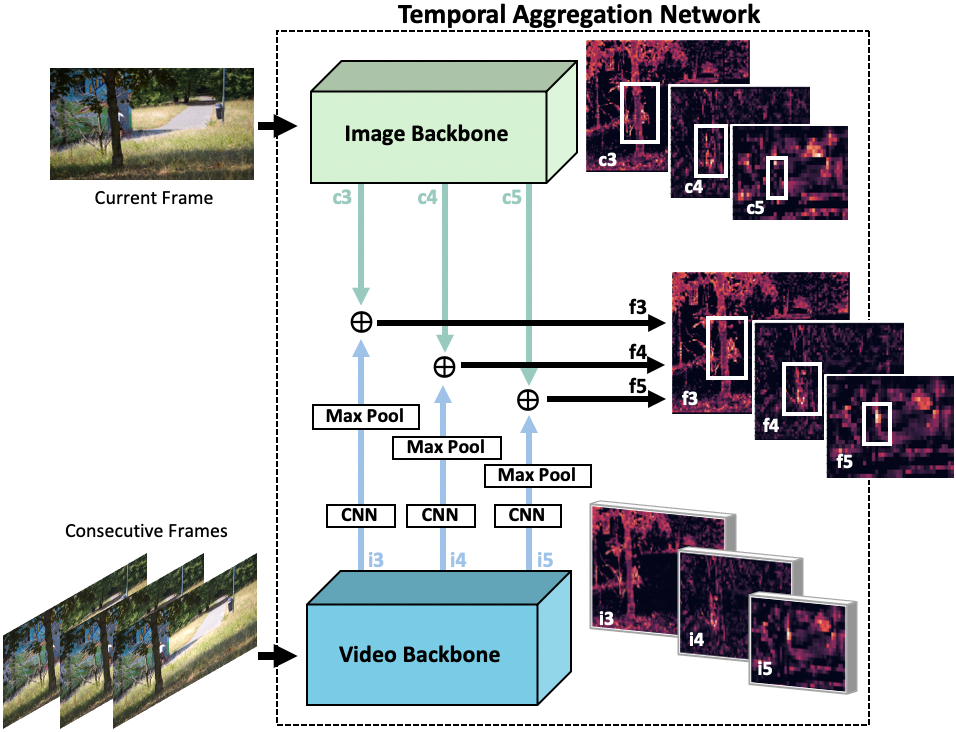}
\end{center}
    \caption{Illustration of Temporal Aggregation Network. Though in c3, c4, c5 from Image Backbone the activation around the target object is weak, after aggregation by Video Backbone, the final output f3, f4, f5 can receive proper activation around the target.}
    \vspace*{-5pt}
\label{fig:fusion}
\end{figure}

Figure~\ref{fig:fusion} illustrates the Temporal Aggregation Network (TAN) with  details to be described in the following.

Many methods in object tracking have exploited temporal information to facilitate tracking.
However, most of them require optical flows to obtain pixel-wise alignment~\cite{zhu2017dff, zhu2017fgfa} which can be quite computationally extensive.
Inspired by the recent progress in large-scale action recognition benchmarks~\cite{carreira2017i3d, gu2018ava}, we propose a novel Temporal Aggregation Network to incorporate backbones in image classification and video recognition, which align well with the tracking-by-detection model as well as detection in image and tracking in video.

\textbf{Image Backbone}
Most object detection methods have exploited ImageNet~\cite{deng2009imagenet} pretrained backbone for faster convergence as studied in~\cite{he2018rethinking}.
We use ResNet~\cite{he2016resnet} as the standard setting similarly done in other detection approaches.
We denote the outputs of stage 3, 4, 5 as c3, c4, c5 respectively.
The input of Image Backbone is the key frame where objects are to be detected.

\textbf{Video Backbone}
Inspired by recent progress in Human Action Recognition~\cite{carreira2017i3d, wang2018nonlocal, feichtenhofer2018slowfast},
we believe 3D convolution is effective in utilizing temporal information across consecutive frames,
although no previous works have attempted to utilize this technique to tackle VOS task.
Note that while optical flow approaches in~\cite{luiten2018premvos,xiao2018jointreid} utilize explicit correspondence,
3D convolution network can directly learn temporal patterns from an RGB stream.
According to~\cite{carreira2017i3d}, I3D performs best against other methods such as CNN+LSTM counterpart.
Therefore, we adopt I3D in our video backbone and at the same time, similar to~\cite{wang2018nonlocal}, $\text{I3D}_{3\times1\times1}$ and $\text{I3D}_{1\times3\times3}$ are used in order to reduce the computation cost of 3D convolution.
We denote the outputs of stage 3, 4, 5 as i3, i4, i5 respectively.
The previous $\alpha$ frames and current key frame are concatenated along the temporal axis resulting in a 3D tensor of size $(\alpha + 1)\times H \times W$ which is then fed into the video backbone.

\textbf{Temporal Aggregation Fusion}
Inspired by SlowFast~\cite{feichtenhofer2018slowfast}, the separation and fusion of fast/slow video stream is a good practice to make use of  temporal information while preserving key frame feature.
As conv-add style in~\cite{feichtenhofer2018slowfast}, the new feature map $\text{f}_i$ is computed by $\text{c}_i + \mathcal{N}(\text{i}_i)$ for stages 3, 4, 5, where $\mathcal{N}$ denotes a small CNN to fuse features with Max Pooling to compress information along the temporal dimension so that the two branches have the same size at each stage.
$\text{f}_i$ will be used in the subsequent detection networks.
Unlike RNN based methods~\cite{xiao2018jointreid, sadeghian2017tracking} which require sophisticated training schedule, our network is feed-forward and can be jointly trained.
As shown in Figure~\ref{fig:tan} in the experimental section, by incorporating temporal information, our system can better handle occlusion.

\subsection{Dynamic Time-evolving Template Matching}
\begin{figure}[]
\begin{center}
   \includegraphics[width=1.0\linewidth]{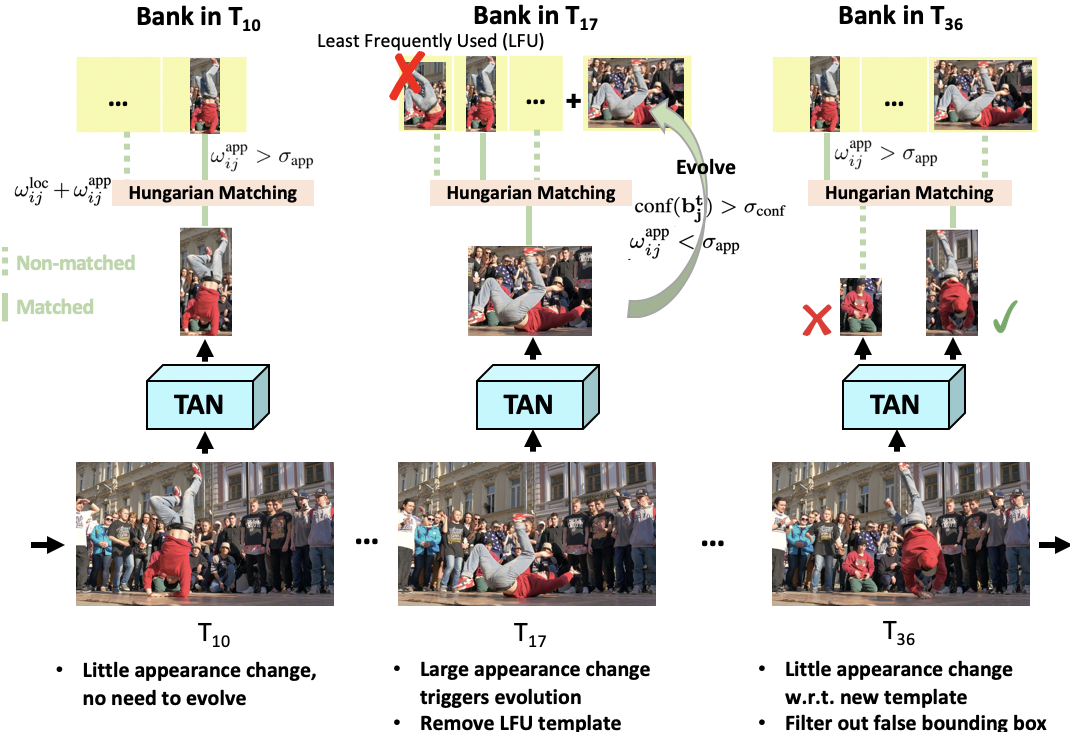}
\end{center}
    \caption{This figure demonstrates an effective case of our DTTM on handling deformation problem. For visualization here we use image to replace feature vector after TAN stage.}
    \vspace*{-15pt}
\label{fig:DTTM}
\end{figure}

Obviously the naive IoU based tracking algorithm is not robust to large camera movement or object deformation.
As pointed out in~\cite{bergmann2019tracking, luiten2018premvos}, re-identification is also essential to object tracking.
Thus we propose the novel Dynamic Time-evolving Template Matching (DTTM) to tackle the re-identification problem. Figure~\ref{fig:DTTM} summarizes the method.

Instead of cropping bounding box regions from RGB image space, we prefer the feature produced by the backbone since it is encoded with high-level semantic meaning. Let
$\mathcal{A}({\bf{b}}^t)$ denote the appearance feature in bounding box $\bf{b}$ extracted from the backbone feature map of frame $t$ which is a high dimensional vector. The appearance similarity weight term is defined as
\begin{equation}\label{eqn:weight-app}
{\omega^{\text{app}}_{ij}} = \frac{\mathcal{A}({\bf{t}}_i) \cdot \mathcal{A}({\bf{b}}^t_j)}{\norm{\mathcal{A}({\bf{t}}_i)} \cdot \norm{\mathcal{A}({\bf{b}}^t_j)}}
\end{equation}
As pointed out in~\cite{sadeghian2017tracking, xu2019spatial}, location cue and appearance cue are essential features in multiple object tracking.
Therefore the bipartite graph is constructed using $\omega_{ij} = \omega^{\text{loc}}_{ij} + \omega^{\text{app}}_{ij}$ in order to take both cues into account, where $\omega_{ij}^{\text{loc}}$ has been defined in Section~\ref{section: tbd}.

In ideal scenarios, the learned appearance feature from the first frame is sufficient for tracking the rest.
However, a constant template clearly does not work in practice especially in long-term tracking, since the target objects may undertake many appearance changes due to deformation, occlusion, etc.
So a proper design for \textbf{time-evolving} template is essential, which leads to our online updating template appearance features during the tracking progress.

Moving average is one of the widely adopted approaches for template update across temporal axis.
The single hyper parameter $\mathit{momentum}$ controls the proportion of features to be updated or preserved,
which is extremely sensitive to the frame rates of different videos.
The static averaging process is also obviously sub-optimal due to accumulation error and feature blurring.
We propose to update the template feature more \textbf{dynamically} in a discrete manner.

\vspace{1mm}
{\bf DTTM}
 Full algorithm detailing our matching strategy in pseudocodes can be found in the supplementary material.
Initially, the template bank for each target $i$ is constructed as $\mathit{bank}_i = \{{\bf{t}}_i\}$.
The matching result is obtained by performing linear assignment between ${\bf{t}}_i \in \cup_{i=1}^{N}  \mathit{bank}_i$ and ${\bf{b}}^t_{j} \in \{{\bf{b}}^t_{j}\}_{j=1}^{N_t}$ at each frame $t$ by computing $\omega^{\text{loc}}_{ij} + \omega^{\text{app}}_{ij}$.
Given foreground confidence score $\text{conf}({\bf{b}}^t_j)$ greater than threshold $\sigma_{\text{conf}}$, when ${\bf{t}}_i$ and ${\bf{b}}^t_j$ is a match but their appearances are not similar $\omega^{\text{app}}_{ij} < \sigma_{\text{app}}$ , which indicates a drastic appearance change and thus a very likely situation of losing track in the upcoming frames,
our DTTM will initialize a new template from the latest matched detection ${\bf{b}}_j^t$ for the target object ${\bf{t}}_k$ and add it into corresponding feature bank, which results in an extended template bank for the future assignment.

To avoid overflow, the least frequently used template will be removed from the $\mathit{\mathit{bank}}_i$ when the number of templates is larger than some threshold.  
Both the initial target feature and updated feature can be considered as templates during later matching process.
Our system can thus detect potential accumulation error early and thus prevent its adverse effect by updating with high confidence new template feature.
 Figure~\ref{fig:DTTM} illustrates how DTTM can effectively address the deformation problem in VOS,
 where not only the latest but also previous templates are considered so that abrupt deformation can be well handled.
 Note that this matching mechanism is entirely online and time-evolving unlike~\cite{luiten2018premvos}, namely, in our DTTM no future information is required and the template bank keeps evolving over time.

\section{Experiments}

\subsection{Dataset and Evaluation Metrics}
We use DAVIS benchmarks~\cite{perazzi2016davis, pont2017davis, caelles2018davis} to evaluate our method.
These benchmarks are challenging, with large variety in scenes, multiple heterogeneous objects and occlusion, and are
widely used in evaluating video object segmentation methods.

\textbf{DAVIS 2016}
DAVIS 2016 comprises of a total of 50 sequences, 3455 annotated frames, all of which were captured at 24fps and Full HD 1080p spatial resolution.
Since computational complexity is a major bottleneck in video processing, each sequence has a short temporal extent (about 2--4 seconds) while including all major challenges typically found in longer video sequences

\textbf{DAVIS 2017}
After releasing DAVIS 2016~\cite{perazzi2016davis}, several strong methods have been proposed making the top performance on DAVIS 2016 saturated.
DAVIS 2017 was released which is an extension of the former version and a more challenging benchmark.
Overall, the new dataset consists of 150 sequences, totaling 10459 annotated frames and 376 objects.
One major difference from DAVIS 2016 is that in DAVIS 2017, multiple object tracking is introduced in video object segmentation.
In DAVIS 2016, only single objects are annotated for each frame while
in DAVIS 2017, multiple object masks in the same frame are annotated,
which makes the task more challenging as complex interaction among multiple target objects can cause occlusion and thus change of topology and appearance.  Moreover, as Figure~\ref{fig:teaser} shows, many target objects are in the same category and have very similar appearance.
Another important challenge in DAVIS 2017 is that  target object category in one video can become background in another video, thus demanding high discriminative ability of the tracker to adapt to different target objects given only the first frame annotation.
Note that DAVIS 2017 is also the dataset for DAVIS 2019 Challenge, which we will use in the following unless otherwise specified.


\subsection{Implementation Details}

We use Faster R-CNN~\cite{ren2015faster} as the detector and ResNet-50~\cite{he2016resnet} as our default backbone unless otherwise specified.
The spatial branch backbone is initialized with weights given by ImageNet~\cite{deng2009imagenet} classification and the detector is pretrained on COCO~\cite{lin2014coco} as done in~\cite{luiten2018premvos}.
As for the temporal aggregation branch, we use 8-frame input I3D baseline provided by Nonlocal Network~\cite{wang2018nonlocal}.
In order to incorporate multiple scales during tracking,
feature pyramid network (FPN)~\cite{lin2017fpn} is adopted in the backbone to merge high-level semantic information from deeper coarse feature map with  shallow fine-grained feature map.
Following~\cite{lin2017fpn}, the RPN anchors span 5 scales (feature is extracted from different levels of FPN outputs) and 3 aspect ratios [0.5, 1.0, 2.0].
Since category classification is unavailable in the DAVIS dataset, the last linear classification layer of R-CNN is replaced by a simple fully connected layer for foreground/background classification.
This class agnostic R-CNN is more suitable for semi-supervised setting and more general for real applications.
Vanilla Fully Convolution Network~\cite{long2015fcn} is adopted as the segmentation head as default.
As in Fast R-CNN~\cite{girshick2015fast}, an RoI is considered positive if its IoU with the ground-truth box is at least 0.5 and negative otherwise.
The segmentation is only performed on positive RoIs during both training and testing.

\begin{figure}[t]
\begin{center}
   \includegraphics[width=1.0\linewidth]{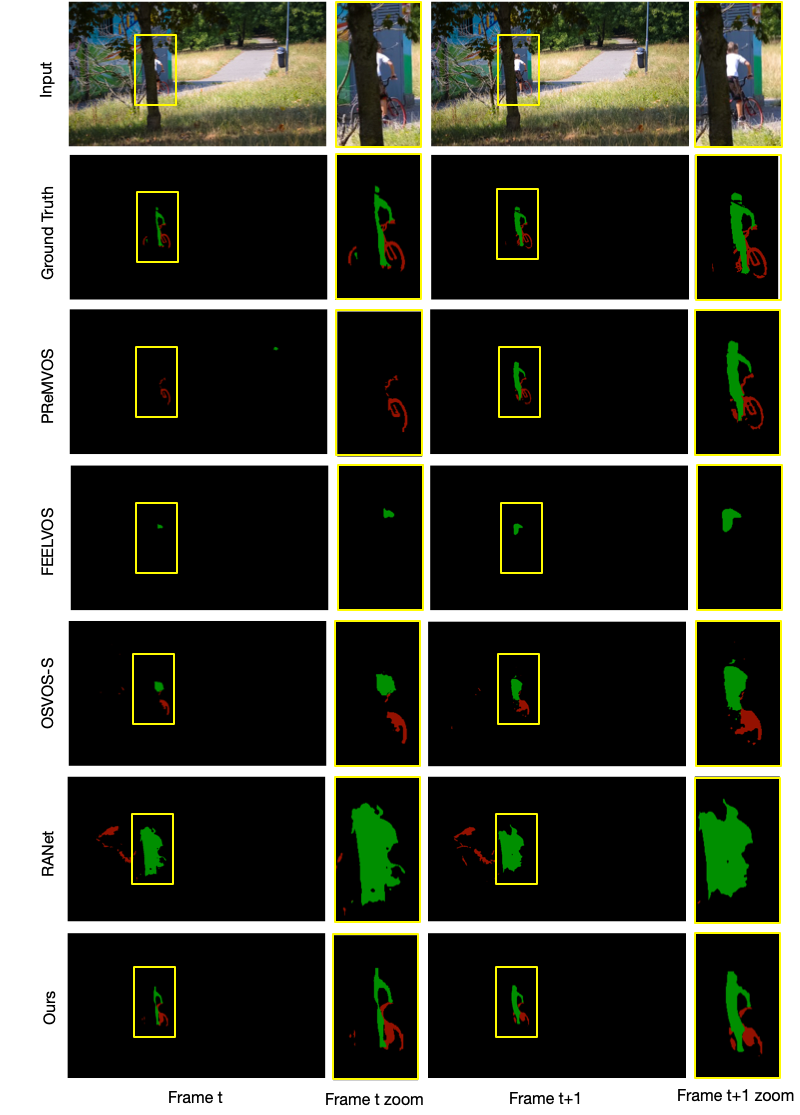}
\end{center}
    \caption{{\bf Disocclusion of man riding bike}. Results of Temporal Aggregation Network compared with Ground Truth, PReMVOS~\cite{luiten2018premvos}, FEELVOS~\cite{voigtlaender2019feelvos}, OSVOS-S~\cite{caelles2017osvos} and RANet~\cite{wang2019ranet}.}
\label{fig:tan}
\end{figure}

\textbf{Training}
To overcome the domain gap between the pretrained dataset and DAVIS, we first train the entire detection network on DAVIS training set.
The  input  images  are  resized  such  that  their  shorter  side  is  800p which is similarly done in~\cite{ren2015faster} unless otherwise specified.
We train on 8 GPUs with 2 images per GPU (effective mini batch size of 16).
All layers of backbone except for c1 and c2 of backbone are jointly fine-tuned with detection annotation.
Due to the limited batch size, all batch normalization layers are frozen so that the statistics of mean and variance are unchanged during training.
Unlike stage-wise training with respect to RPN in~\cite{he2017mask}, end-to-end  training  as in~\cite{lin2017fpn} is adopted in our implementation which yields better results.
All models are trained for 100k iterations using synchronized SGD with a weight decay of 0.0001 and momentum of 0.9.
The learning rate is initialized to 0.002, which decays by a factor of 10 after 70k iterations and 90k iterations.
Other choices of hyper-parameters also follow the setting in~\cite{ren2015faster}.

\textbf{Inference}
At test time, we mostly follow the setting of Faster-RCNN in \cite{lin2017fpn}.
The box prediction is done on RPN proposals,
followed by non-maximal suppression \cite{ross2010dpm}.
The segmentation head is then applied to the detected bounding boxes above threshold $\sigma_{\text{conf}}$ only.
The instance segmentation results are merged into a single segmentation map by  ranking  corresponding bounding box confidence scores.

\subsection{Ablation Study}
The ablation study is performed on the DAVIS 2017 dataset.
The outputs are resized to 480p for evaluation.

\vspace{-15pt}
\subsubsection{Temporal Aggregation Network}
\vspace{-5pt}

We report the standard object detection and instance segmentation COCO metrics including AP, AP50, AP75 for both bounding boxes and segmentation masks. Figure~\ref{fig:tan} shows that our Temporal Aggregation Network is more robust to occlusion compared with other methods.

\begin{table}[]
  \small
    \centering
    \addtolength{\tabcolsep}{-5pt}
    \scalebox{0.95}{%
\begin{tabular}{c|c|ccc|ccc}
\Xhline{1.0pt}
res & TAN & AP${^\text{bbox}}$ & AP$^\text{bbox}_\text{50}$ & AP$^\text{bbox}_{75}$&AP$^\text{mask}$&AP$^\text{mask}_\text{50}$&AP$^\text{mask}_\text{75}$ \\
\hline
800 &&$45.3$&$66.5$&$51.1$&$35.4$&$59.4$&$36.9$  \\
800 &\checkmark&$\mathbf{46.3_{\uparrow1.0}}$&$\mathbf{70.9_{\uparrow4.4}}$&$\mathbf{52.3_{\uparrow1.2}}$&$\mathbf{36.7_{\uparrow1.3}}$&$\mathbf{60.9_{\uparrow1.5}}$&$\mathbf{38.1_{\uparrow1.2}}$  \\
\hline
600 &&$45.0$&$66.5$&$51.0$&$35.3$&$59.0$&$37.2$ \\
600 &\checkmark&$\mathbf{46.3_{\uparrow1.3}}$&$\mathbf{71.0_{\uparrow4.5}}$&$\mathbf{52.5_{\uparrow1.5}}$&$\mathbf{36.2_{\uparrow0.9}}$&$\mathbf{61.0_{\uparrow2.0}}$&$\mathbf{38.0_{\uparrow0.8}}$\\
\hline
400 &&$44.5$&$66.6$&$50.9$&$35.2$&$58.9$&$\mathbf{37.5}$  \\
400 &\checkmark&$\mathbf{46.6_{\uparrow2.1}}$&$\mathbf{71.2_{\uparrow4.6}}$&$\mathbf{52.9_{\uparrow2.0}}$&$\mathbf{35.8_{\uparrow0.6}}$&$\mathbf{62.9_{\uparrow4.0}}$&$36.9_{\downarrow0.6}$ \\
\Xhline{1.0pt}
\end{tabular}}
\caption{Results of Temporal Aggregation Network (TAN) on DAVIS validation set.
This table presents results under different input resolutions when comparing with the baseline (i.e. no TAN).}
\vspace{-1em}
\label{table:ablation-tan-davis}
\end{table}

\textbf{Different Input Resolutions}
Although high-resolution input evidently benefits the localization ability of a network, it is not always applicable due to  hardware constraints. Table~\ref{table:ablation-tan-davis} shows that
even though the computation cost increases quadratically when input resolution increases, the performance gain is only marginal and saturates very soon (e.g.\ first column, AP${^\text{bbox/mask}}$ 45.3/35.4 at 800 res.\ versus AP${^\text{bbox/mask}}$ 44.5/35.2 at 400 res.\ ).
By aggregating temporal information, even low-resolution input can achieve better performance (e.g.\ first column, AP${^\text{bbox/mask}}$ 45.3/35.4 at 800 res.\ versus AP${^\text{bbox/mask}}$ 46.6/35.8 at 400 res.\ with TAN). 
It is also worth noting that the baseline method performance drops significantly under low-resolution setting,
while from different input resolutions TAN achieves consistently higher performance,
which indicates that effectively aggregating temporal information across frames can complement loss of resolution due to compression to some extent.

\textbf{More Challenging Dataset}
Since the number of target object in DAVIS dataset is quite limited (1 in DAVIS 2016, 1 to 3 in DAVIS 2017),
we also demonstrate the effectiveness of our temporal aggregation network on ImageNet VID dataset~\cite{deng2009imagenet}
which is larger (3962 training sequences, 555 validation sequences) and more complicated (1 to 10 target objects from 30 categories).
Table~\ref{table:ablation-tan-vid} presents our quantitative results.
Our method's performance is on par with popular optical flow feature alignment methods with stronger backbones.
In addition, using continuous frames as input can boost up performance by 2 points compared with duplicate frames,
which again demonstrates that temporal information is effectively utilized by our proposed Temporal Aggregation Network.

\begin{table}[]
  \small
    \centering
    \addtolength{\tabcolsep}{-3pt}
\begin{tabular}{c|c|c|c|c}
\Xhline{1.0pt}
method & optical flow & backbone & input &AP$^\text{bbox}_\text{50}$ \\
\hline
DFF~\cite{zhu2017dff} &\textbf{\checkmark}&ResNet-101&Continuous& $73.0$   \\
FGFA~\cite{zhu2017fgfa} &\textbf{\checkmark}&ResNet-101&Continuous& $76.8$  \\
\hline
\textbf{Ours} &  &ResNet-50&Duplicate&  $76.0$  \\
\textbf{Ours} &  &ResNet-50&Continuous&  $\mathbf{78.2_{\uparrow2.2}}$ \\
\Xhline{1.0pt}
\end{tabular}
\caption{Results of Temporal Aggregation Network (TAN) on ImageNet VID validation set.
Duplicate denotes simply stacking the same frame to fit the network input size, while continuous means both current and neighbor frames are fed into TAN.
}
\vspace{-1.5em}
\label{table:ablation-tan-vid}
\end{table}

\vspace{-0.5em}
\subsubsection{Dynamic Time-evolving Template Matching}

\begin{figure}[t]
\begin{center}
   \includegraphics[width=1.0\linewidth]{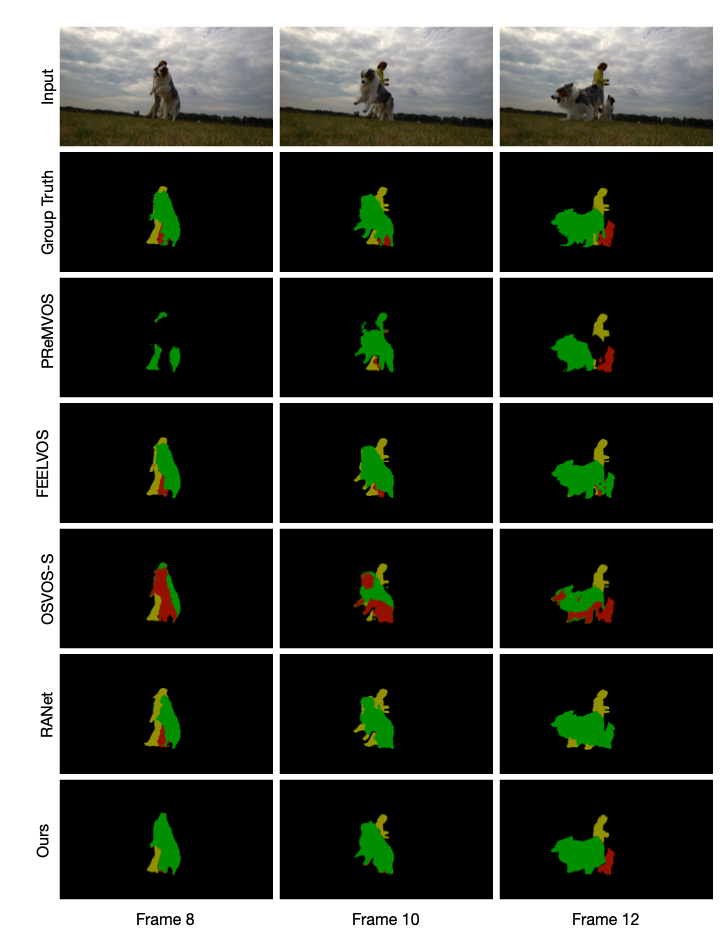}
\end{center}
    \caption{{\bf Large deformation (foreground dog) with disocclusion of a similar but different instance (background dog)}. Results of Dynamic Time-evolving Template Matching compared with Ground Truth, PReMVOS~\cite{luiten2018premvos}, FEELVOS~\cite{voigtlaender2019feelvos}, OSVOS-S~\cite{caelles2017osvos} and RANet~\cite{wang2019ranet}.}
\label{fig:dttm-vis}
\end{figure}

Besides no matching and IoU matching, we also compare our dynamic template matching with naive moving average with momentum $\mathit{mnt}$, in which matching is based on $\mathcal{A}({\bf{t}}_i)=(1-\mathit{mnt})  \mathcal{A}({\bf{t}}_i) + \mathit{mnt} \mathcal{A}({\bf{b}}_j)$ where ${\bf{t}}_i$ and ${\bf{b}}_j$ is a match.
As Table~\ref{table:ablation-dtt} shows, by setting $\theta_{\text{conf}}=0.5$ and sliding $\theta_{\text{app}}$, our dynamic template matching outperforms the baseline by a large margin.
Note that for the DTTM module, if the threshold becomes too low (e.g. $0.3$), it may introduce misleading templates.
On the other hand, if the threshold becomes too high (e.g. $0.7$), it may filter out useful templates. Either case will lead to performance drop.
Note that even though moving average can also improve the performance, it is worse than DTTM in general.
We argue that due to the high frame rate of DAVIS dataset, it is hard for naive moving average to model appearance changes which makes it more prone to accumulation error.

Qualitative comparison with leading approaches are provided in Figure~\ref{fig:dttm-vis}.
Multiple major tracking challenges are presented in this example, such as topology change, occlusion, similar texture pattern and semantic. Embedding and segmentation based approaches like \cite{voigtlaender2019feelvos, caelles2017osvos, wang2019ranet} failed to distinguish two dogs and even produce shattered segmentation.
Moreover \cite{luiten2018premvos} failed to incorporate drastic topology (pose) change of the jumping dog resulting in false tracking, while DTTM can easily delineate targets in similar appearance and keep tracking of rapidly deforming objects.

\begin{table}[]
\small
\centering
\addtolength{\tabcolsep}{-1pt}
\begin{tabular}{@{\hspace{0mm}}c|c|c@{\hspace{2mm}}c@{\hspace{2mm}}c}
\Xhline{1.0pt}
method & setting & $\mathcal{J}$\&$\mathcal{F}$-Mean & $\mathcal{J}$-Mean & $\mathcal{F}$-Mean \\
\Xhline{1.0pt}
IoU match & &$69.2$&$65.8$&$72.6$ \\
\hline
moving avg & $\mathit{mnt}=0.2$ & $70.1_{\uparrow0.9}$ & $66.8_{\uparrow1.0}$ & $73.4_{\uparrow0.8}$ \\
moving avg & $\mathit{mnt}=0.3$ & $\mathbf{70.8_{\uparrow1.6}}$ & $\mathbf{67.5_{\uparrow1.7}}$ & $\mathbf{74.1_{\uparrow1.5}}$ \\
moving avg & $\mathit{mnt}=0.5$ & $69.4_{\uparrow0.2}$ & $66.1_{\uparrow0.3}$ & $72.8_{\uparrow0.2}$\\
\hline
DTTM & $\theta_{\text{app}}=0.3$ &$70.5_{\uparrow1.3}$&$67.3_{\uparrow1.5}$&$73.7_{\uparrow1.1}$\\
DTTM & $\theta_{\text{app}}=0.5$ &$\mathbf{71.7_{\uparrow2.5}}$&$\mathbf{68.5_{\uparrow2.7}}$&$\mathbf{74.9_{\uparrow2.3}}$\\
DTTM & $\theta_{\text{app}}=0.7$ &$70.6_{\uparrow1.4}$&$67.4_{\uparrow1.6}$&$73.9_{\uparrow1.3}$\\
\Xhline{1.0pt}
\end{tabular}
\caption{Results of moving average and Dynamic Time-evolving Template Matching (DTTM) on DAVIS validation set.}
    \vspace{-1.5em}
\label{table:ablation-dtt}
\end{table}

\begin{figure*}[t]
\begin{center}
   \includegraphics[width=0.9\linewidth]{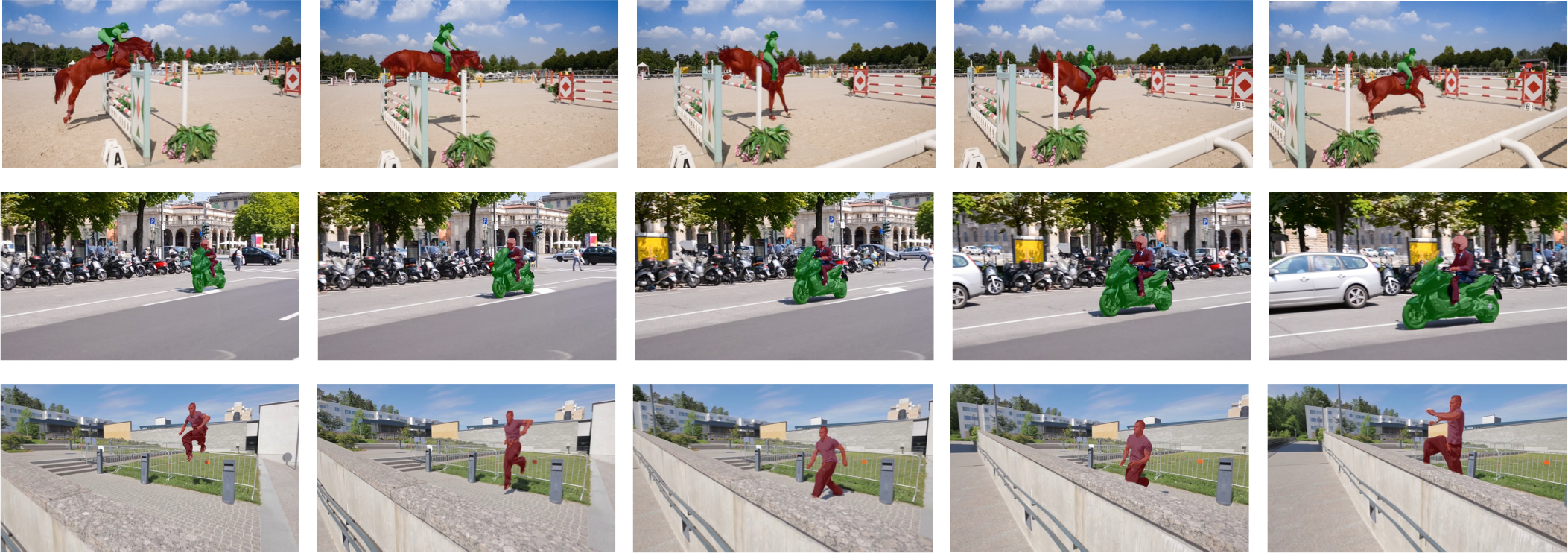}
\end{center}
   \caption{Qualitative results on DAVIS validation set.
   Several challenging cases are presented, such as occlusion, deformation, zoom in/out, to demonstrate robustness of our method.}
\label{fig:big-teaser}
\vspace{-0.2in}
\end{figure*}

\vspace{-10pt}
\subsubsection{Segmentation Head}

\begin{table}[]
\small
\centering
\addtolength{\tabcolsep}{-1pt}
\begin{tabular}{c|c|ccc}
\Xhline{1.0pt}
head & skip connection & $\mathcal{J}$\&$\mathcal{F}$-Mean & $\mathcal{J}$-Mean & $\mathcal{F}$-Mean \\
\Xhline{1.0pt}
FCN &&$67.9$&$64.2$&$71.5$  \\
\hline
FCN &\checkmark \textit{cascade}&$68.8_{\uparrow0.9}$&$65.4_{\uparrow1.2}$&$72.2_{\uparrow0.7}$  \\
FCN &\checkmark \textit{join}&$\mathbf{69.2_{\uparrow1.3}}$&$\mathbf{65.8_{\uparrow1.6}}$&$\mathbf{72.6_{\uparrow1.1}}$\\
\Xhline{1.0pt}
\end{tabular}
\caption{Results of different types of segmentation heads on DAVIS validation set. These results are produced without using TAN and DTTM.}
\label{table:ablation-segmentation-head}
\vspace{-15pt}
\end{table}

Different off-the-shelf design choices of segmentation head have also been investigated.
We deploy two types of skip connection, \textit{cascade} and \textit{join}, where \textit{cascade} sequentially adds the output of each layer to the input of next layer as residue, and \textit{join} simply aggregates all layer outputs by addition as final output.
Table~\ref{table:ablation-segmentation-head} shows the results of different segmentation heads.
Note the simple skip connection has significant improvement over the vanilla FCN counterpart,
which indicates that our method can be further benefited from advanced segmentation techniques.

\vspace{-0.5em}
\subsection{Results on Benchmark}
\vspace{-0.5em}
Table~\ref{table:comparisons-stoa} tabulates the quantitative results in comparison with  leading methods with qualitative results presented in Figure~\ref{fig:big-teaser}.
We achieve state-of-the-art single model results in both speed and accuracy on DAVIS Benchmark with stronger backbone ResNeXt-101~\cite{xie2017resnext} and deformable convolution~\cite{dai2017deformable},
both contribute to further improving  overall performance, which again justifies the advantage of our simple and easily extensible pipeline.
Despite DAVIS dataset, many leading methods exploit YouTube-VOS~\cite{xu2018youtube} as pretraining. In Table~\ref{table:comparisons-stm} we demonstrated that our method achieved higher performance without computational demanding YouTube-VOS pretraining. The speed (t/s) reported in Table~\ref{table:comparisons-stm} considers that there are more than 2 objects on average on DAVIS-2017 dataset. STM~\cite{oh2019stm} needs to handle them individually, while we could detect and track all the targets in one pass. We also reported a reasonable result on YouTube-VOS, $73.5\%$ $\mathcal{J} \& \mathcal{F}$-Mean without carefully selecting hyperparameter. 
\begin{table}
\small
\centering
\addtolength{\tabcolsep}{-3.5pt}
\begin{tabular}{c|c|ccc}
\Xhline{1.0pt}
Method & t/s & $\mathcal{J}$\&$\mathcal{F}$-Mean & $\mathcal{J}$-Mean & $\mathcal{F}$-Mean \\
\Xhline{1.0pt}
PReMVOS~\cite{luiten2018premvos} &$37.6$&$\mathbf{77.8/71.6}$&$\mathbf{73.9/67.5}$&$\mathbf{81.7/75.8}$ \\
OnAVOS~\cite{voigtlaender2017online}  &$26$&$63.6/52.8$&$61.0/49.9$&$66.1/55.7$ \\
FAVOS~\cite{cheng2018favos}  &$1.2$&$58.2/43.6$&$54.6/42.9$&$61.8/44.2$ \\
VideoMatch~\cite{hu2018videomatch} &$0.35$&$62.4/-$&$56.5/-$&$68.2/-$ \\
FEELVOS~\cite{voigtlaender2019feelvos} &$0.51$&$71.5/57.8$&$69.1/55.1$&$74.0/60.4$ \\
OSMN~\cite{yang2018efficient} &$\mathbf{0.28}$&$54.8/41.3$&$52.5/37.7$&$57.1/44.9$ \\
RGMP~\cite{wug2018fast}    &$\mathbf{0.28}$&$66.7/52.8$&$64.8/51.3$&$68.6/54.4$ \\
\hline
\textbf{Ours} &$\mathbf{0.14}$&$\mathbf{75.9/65.4}$&$\mathbf{72.3/61.3}$&$\mathbf{79.4/70.3}$\\
\Xhline{1.0pt}
\end{tabular}
\caption{Comparisons with state-of-the-art methods on the \textbf{validation/test-dev set} of DAVIS 2019 Challenge. t/s denotes running time per frame in seconds.
The table demonstrates that our method achieves state-of-the-art performance in both speed and accuracy.}
\label{table:comparisons-stoa}
\vspace{-0.2in}
\end{table}

\begin{table}[!htb]
\small
\centering
\addtolength{\tabcolsep}{-3.5pt}
\begin{tabular}{c|c|c|ccc}
\Xhline{1.0pt}
Method & t/s & YV & $\mathcal{J}$\&$\mathcal{F}$-Mean & $\mathcal{J}$-Mean & $\mathcal{F}$-Mean \\
\Xhline{1.0pt}
\multirow{2}{*}{FEELVOS~\cite{voigtlaender2019feelvos}} & \multirow{2}{*}{0.51} &  & 69.1/54.4 & 65.9/51.2 & 72.3/57.5 \\
 &  & \checkmark & 71.5/57.8 & 69.1/55.2 & 74.0/60.5 \\
\hline
\multirow{2}{*}{STM~\cite{oh2019stm}} & \multirow{2}{*}{0.32} &  & 71.6/- & 69.2/- & 74.0/- \\
 &  & \checkmark & 81.7/72.2 & 79.2/69.3 & 84.3/75.2 \\
 \hline
\textbf{Ours} & 0.14 &  & 75.9/65.4 & 72.3/61.3 & 79.4/70.3 \\
\Xhline{1.0pt}
\end{tabular}
\caption{Comparisons with state-of-the-art methods on the \textbf{validation/test-dev set} of DAVIS-2017. YV denotes whether YouTube-VOS is used during training.}
    \vspace*{-15pt}
\label{table:comparisons-stm}
\end{table}

\vspace{-0.5em}
\section{Conclusion}
\vspace{-0.5em}
Many leading VOS methods are overly complicated utilizing computationally-heavy modules or highly-engineered pipelines leading to limited practical usage. In this paper, we design a new and strong baseline that simultaneously achieves state-of-the-art speed and accuracy, by integrating the “tracking-by-detection” model into VOS, since VOS can be naturally decoupled into image semantic segmentation and video object tracking.
With this design, our method is easy to extend because ongoing advancement in object tracking can further improve our method in future.
With the introduction of multiple object segmentation in the DAVIS 2017 challenge, most leading methods at the time supporting multiple object segmentation required extra modification,
whereas our method handles multiple object segmentation in one forward pass.
On top of our design, we propose the novel Temporal Aggregation Network (TAN) and Dynamic Time-evolving Template Matching (DTTM), and their effectiveness have been demonstrated experimentally.
Without bells and whistles, our method achieves a new state-of-the-art result on DAVIS benchmark for VOS.
We hope our fast, practical and easy to extend pipeline will serve as a new baseline for future development targeting at higher efficiency,  accuracy and  scalability in VOS.

{\small
\bibliographystyle{ieee_fullname}
\bibliography{vosbib}
}

\end{document}